\documentclass[10pt,twocolumn,letterpaper]{article}

\usepackage{cvpr}              

\usepackage{graphicx}
\usepackage{amsmath}
\usepackage{amssymb}
\usepackage{booktabs}

\usepackage[pagebackref,breaklinks,colorlinks]{hyperref}

\usepackage[capitalize]{cleveref}
\crefname{section}{Sec.}{Secs.}
\Crefname{section}{Section}{Sections}
\Crefname{table}{Table}{Tables}
\crefname{table}{Tab.}{Tabs.}

\def\cvprPaperID{*****} 
\def\confName{CVPR}
\def\confYear{2022}

\begin{document}

\title{Unsupervised Learning of Depth, Camera Pose and Optical Flow from Monocular Video}

\author{Dipan Mandal\\
Senior Member, IEEE\\
{\tt\small dipan.mandal@gmail.com}
\and
Abhilash Jain\\
Processor Architecture Research Lab, Intel Corporation\\
{\tt\small abhilash.jain@intel.com}
}

\maketitle

\begin{abstract}
We propose DFPNet -- an unsupervised, joint learning system for monocular \underline{D}epth, Optical \underline{F}low and egomotion (Camera \underline{P}ose) estimation from monocular image sequences. Due to the nature of 3D scene geometry these three components are coupled. We leverage this fact to jointly train all the three components in an end-to-end manner. A single composite loss function - which involves image reconstruction-based for depth \& optical flow, bi-directional consistency checks and smoothness loss components - is used to train the network. Using hyperparameter tuning, we are able to reduce the model size to \textnormal{less than 5\%} ($\sim$8.4M parameters) of state-of-the-art DFP models. Evaluation on KITTI and Cityscapes driving datasets reveals that our model achieves results comparable to state-of-the-art in all of the three tasks, even with the significantly smaller model size.
\end{abstract}

\section{Introduction}
\label{sec:intro}

Humans develop the capability of interpreting 3D scene geometry of even an unknown scene from their vast experience of seeing how 3D structures in real world interacts. This capability is remarkably robust exactly in the cases that the ‘conventional methods’ struggle to perform well – for example, modeling luminance variation in indoor vs outdoor scenes, high dynamic range induced pixel saturation, high/low motion of scene objects, understanding motion and viewpoint change induced occlusions, non-ideal optical surfaces and transparencies, disambiguating repetitive structures or low-textured surfaces etc. In fact, supervised learning - applied individually to depth, optical flow and pose estimation problems – does work to some extent. However, especially for these 3D geometry related estimation problems, determining accurate ground truth for supervisions prohibitively complex, slow and more importantly, often inaccurate, ambiguous and sometimes, is simply impossible. Inaccurate supervision can actually do more harm than good – it may lead to wrong optimization (or, ineffective generalization to other scenes types) of a learning framework. Additionally, we argue that a deep learning approach for estimating any one individual task (of depth, flow and pose estimation) inherently learns common characteristics of the scene (e.g. geometric structure of the scene, rigid body dynamics of the scene objects, camera model etc.) and this learning could be reused or generalized to learning other tasks as well.\par

\begin{figure}[t]
  \centering
   \includegraphics[width=1\linewidth]{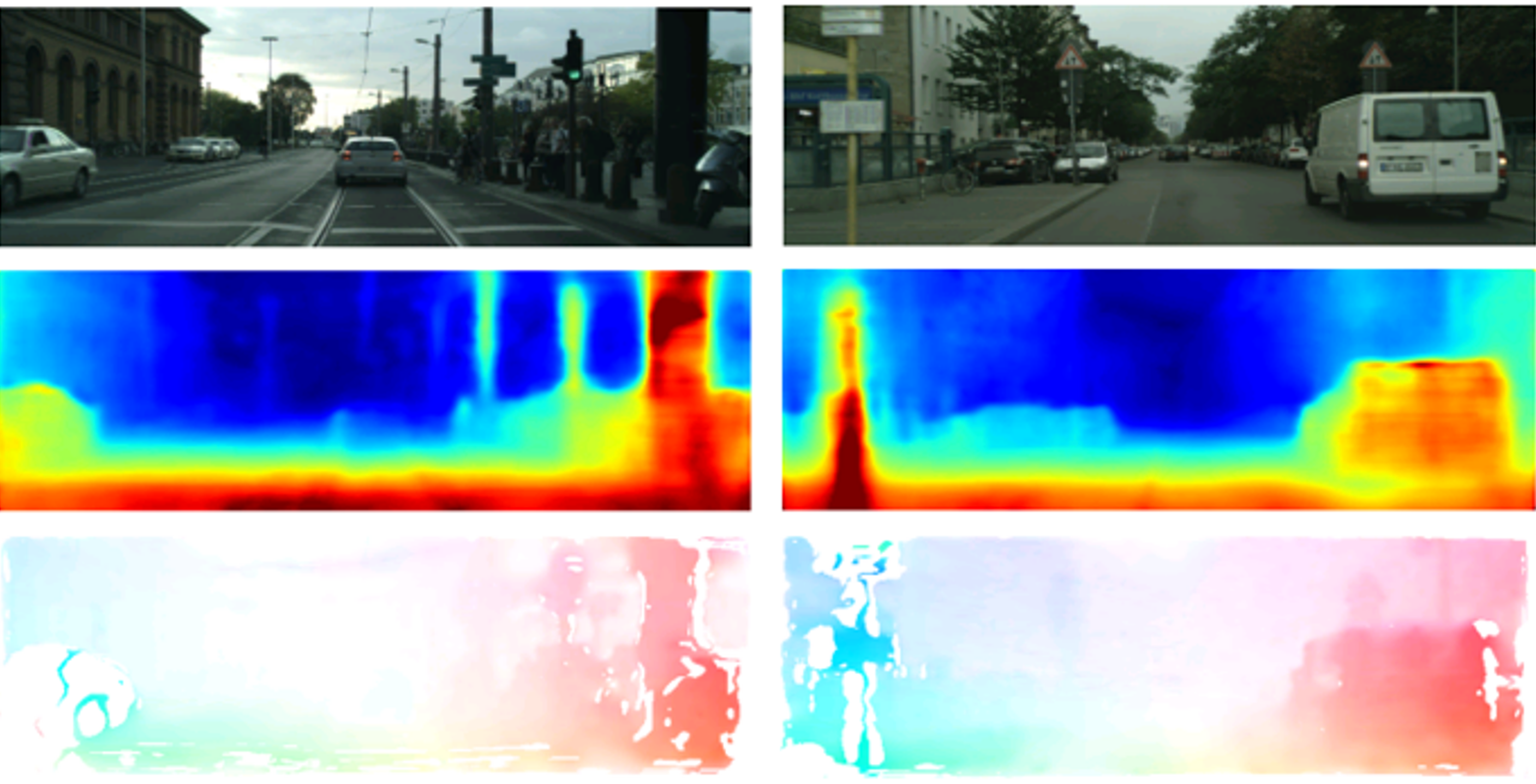}

   \caption{Example predictions of our method on Cityscapes. Top to bottom: input image (image 0 of the sequence), depth map and optical flow.}
   \label{fig:picture1}
\end{figure}

We also note that a monocular (single) camera based depth, pose, pose estimation may result in large savings in system cost, energy and enable smaller form factor system designs compared to stereo (dual) camera based solutions. The advantage amplifies for systems requiring wide field-of-view coverage with multiple cameras (e.g. drones). We also note that a well-trained, multi-task learning algorithm, apart from estimating all quantities simultaneously, can also be applied to each problem individually (e.g. estimate only depth or only optical flow) and with any meaningful combination (e.g. estimate optical flow and camera pose) to suite a system’s capability, hardware configuration, use case specific need, even at runtime. This provides a more efficient way to design a SoC/system with capabilities that can scale based on use cases yet leveraging same core algorithm, hardware engines/IPs and software infrastructure.\par
In this paper, we explore a monocular, self-supervised, jointly learnt model for depth, optical flow and camera pose estimation, which we call DFPNet. As future work, we will undertake development of a common hardware architecture co-optimized for the model. We maintain special focus on keeping the model parameters as low as possible, to enable fast and real-time execution.\par
DFPNet is a multi-head, auto-encoder type deep learnt architecture. It is trained it in an unsupervised fashion on a variety of monocular datasets. The estimated scene depth, optical flow and camera pose can be directly used by several high level, task specific, evolving visual AI tasks (e.g. object tracking, navigation, dynamic obstacle avoidance, motion or path planning, AR/VR scene augmentation) or can be used to build highly robust simultaneous localization and mapping (SLAM) pipeline for long term environment analysis.\par

\section{Related Work}

\subsection{Conventional 3D Scene Understanding}

Understanding a 3D scene has been well-established as a structure-from-motion (SfM) learning problem. This is typically done by using a very large set of unordered images to deduce scene structure and camera motion jointly \cite{furukawa2015multi, hartley2003multiple}. More recent approaches usually start with feature extraction and matching, followed by geometric verification \cite{schonberger2016structure}. Bundle adjustment \cite{triggs1999bundle} is then used for reconstruction of the structure. However, existing methods need accurate feature matching for good 3D scene reconstruction. Without accurate and robust photo-consistency, the performance cannot be guaranteed. Typical failure cases in real-world application involve low textured surfaces, stereo ambiguities, occlusions, etc.\par
Another aspect of understanding 3D scene geometry is scene flow estimation, which involves estimating the dense motion field of a scene from image sequences \cite{vedula1999three}. Top ranked methods on the KITTI benchmark typically involve the joint reasoning of geometry, rigid motion and segmentation \cite{behl2017bounding, vogel20153d}.\par
Several recent methods emphasize the rigid regularities in generic scene flow. Methods like masking out moving objects from a rigid scene \cite{taniai2017fast}, using semantic segmentation to define multiple models of image motion \cite{sevilla2016optical} have been proposed. \cite{wulff2017optical} modified the Plane+Parallax framework with semantic rigid prior learned by a CNN.\par
In our approach, we make use of deep convolution neural networks (CNN) for better exploitation of high level cues, not restricted to a specific scenario. We use a pyramidal-feature extraction backbone to extract features at multiple scales. These features are then processed upon by specific decoders to predict the dense depth, optical flow and relative camera pose in a single pass of the network.

\subsection{Supervised Learning for 3D Scene Understanding}
Recently, deep learning has been successfully employed in many tasks of 3D scene understanding, including depth, optical flow, pose estimation, etc. These advances have mostly been due to efficient algorithm development, better hardware for faster training and availability of large data corpus used to train the models.\par
Eigen et al. \cite{eigen2014depth} showed how deep models can estimate single view depth by using a two-scale network. Since such monocular formulation relies majorly on scene priors, many recent methods prefer to use a stereoscopic setup. Mayer et al. \cite{mayer2016large} introduced a correlation layer to mimic traditional stereo matching techniques. Kendall et al. \cite{kendall2017end} proposed 3D convolutions over cost volumes by deep features to better aggregate stereo information.\par
Optical flow learning has seen a similar trend. E. Ilg et al. \cite{ilg2017flownet} made use of a large synthetic data to train a stacked network and achieved results on par with conventional methods. Sun et al. \cite{sun2018pwc} used pyramidal processing, intermediate flow warping, and "cost volume" to achieve impressive optical flow results with a much smaller and faster model compared to it's predecessors. Teed et al. \cite{teed2020raft} employed recurrent transformer architecture and obtained state-of-the-art optical flow results.\par
Apart from these dense pixel prediction problems, camera localization and tracking have also proven to be tractable as a supervised learning task. Kendall et al. \cite{kendall2015posenet} cast the 6-DoF camera pose relocalization problem as a learning task, and extended it upon the foundations of multiview geometry. Oliveira et al. \cite{oliveira2020topometric} demonstrated how to assemble visual odometry and topological localization modules and outperformed traditional learning-free methods. Brahmbhatt et al. \cite{brahmbhatt2018geometry} exploited geometric constraints from a diversity of sensory inputs for improving localization accuracy on a broad scale.

\subsection{Unsupervised Learning 3D Scene Understanding}
To avoid using expensive and difficult to obtain ground truth data, many unsupervised approaches have been proposed to address the 3D understanding tasks. In the absence of ground truth, the core supervision for training is formulated as a view synthesis task based on geometric inferences. Here we briefly review on the most closely related ones.\par
Garg et al. \cite{garg2016unsupervised} proposed an auto-encoder type model for single view depth estimation from a stereo setup. Ren et al. \cite{ren2017unsupervised} and Yu et al. \cite{yu2016back} expanded on the image reconstruction loss together with a spatial smoothness loss for unsupervised optical flow learning. On the other hand, Godard et al. \cite{godard2017unsupervised} exploited geometric consistency constraints in monocular depth estimation with a left-right consistency loss. Zhou et al. \cite{zhou2017unsupervised} coupled the monocular depth and ego-motion learning task to solve the traditional structure-from-motion task. Extending the rigid projective geometry, these methods do not consider the dynamic objects explicitly. Instead, they learn a separate "explainability" mask to account for such motions. Correspondingly, Vijayanarasimhan et al. \cite{vijayanarasimhan2017sfm} learned several object masks and related rigid motion parameters for modelling moving objects.\par
In our approach, we use a combination of some of the aforementioned methods: an auto-encoder type network architecture, image reconstruction-based and smoothness losses, along with forward-backward consistency checks as our objective function. Our key supervision method is based on view synthesis and image reconstruction.

\section{Method}
In this section, we first give an overview of our proposed network DFPNet. Following which, we describe the view synthesis and flow image warping methods used for supervision. Finally, we explain the objective function used.

\subsection{DFPNet Overview}
The architecture of DFPNet is an auto-encoder like deep CNN. The encoder is a pyramidal feature extractor which extracts features at different levels of spatial dimension. At each level, three convolution layers are used to extract features. The obtained feature pyramid is then passed onto three different decoder heads: depth decoder (DD), optical flow decoder (FD) and pose decoder (PD). The input to our network is a sequence of two consecutive images captured from a moving camera ($I_t$ and $I_{t+1}$). Let the image sizes be $H \times W \times 3$. We convert both the images to grayscale and concatenate them to form an input size of $H \times W \times 2$ before feeding into the network. The feature pyramid for the both the images is extracted jointly to save computation time. Later, the two pyramids are separated and used accordingly with the decoder. The output of DFPNet is three separate quantities: a dense depth map ($H \times W \times 1$), a dense optical flow field ($H \times W \times 2$) along with an occlusion mask ($H \times W \times 1$) and a 6-dimensional relative pose vector ($6 \times 1$).\par

\begin{figure}[h]
   \centering
   \includegraphics[width=1\linewidth]{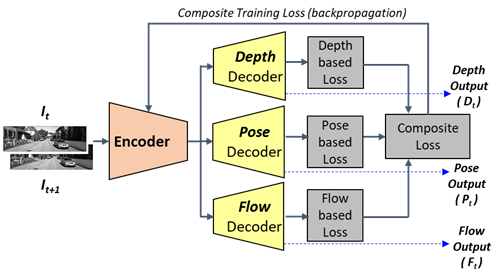}
   \caption{Overview of DFPNet}
   \label{fig:picture2}
\end{figure}

The DD operates on the feature pyramid of $I_t$ from top to bottom: predicting a depth map for the highest pyramid level (and smallest spatial dimension) first. This output is used in the next level, where it is first concatenated with the corresponding features from the pyramid. This concatenated feature set is used to predict a higher resolution depth map. This process is iteratively carried out till the base of the pyramid is reached. This approach, where the features at different levels are concatenated with output at the corresponding level is inspired from the U-Net architecture proposed by Ronneberger et al. \cite{ronneberger2015u}.\par
The PD architecture is a series of convolution layers for feature extraction followed by a global average pooling layer to get a one-dimensional vector. Similar to DD, the pose net also has a U-Net like structure where it iteratively updates the pose from the top to the bottom level. At each level of pose estimation, the features from the corresponding level are used for prediction.
The FD architecture is inspired from PWCNet \cite{sun2018pwc}. One optical flow field is predicted for each level of the feature pyramid and similar to DD and PD, incrementally updated to finally give a flow output of size $H \times W \times 2$. At each level, first a warp is applied to the feature set of $I_t$ using the flow field predicted at the previous level. Next, a cost volume is computed by taking the correlation between the warped $I_t$ and $I_{t+1}$. Optical flow is then predicted using the cost volume and the two feature sets. A final upsampling of the predicted flow is done so that it can be used for warping in the next level. (For more details on the cost volume and flow prediction, please refer \cite{sun2018pwc}). The FD also outputs an occlusion mask which depicts the occluded regions between the two images.\par
Loss values are calculated using view synthesis and image reconstruction (described in the next section) for all the three decoder heads. A final composite loss is calculated by taking a weighted average of the three losses, which is then used for final weight update.\par
For regularization, we use spatial dropout and group normalization throughout the network. The activation used at all the relevant layers is leaky ReLU. Figure \ref{fig:picture2} shows an overview of DFPNet.

\subsection{Supervision Technique}
While formulating the objective function, we assume that the scenes used during training are rigid i.e., the scene appearance change across frames is caused by the camera motion rather than object motion. Although our training dataset contains scenes which are not rigid, our framework is robust to some degree of scene motion.

\subsubsection{Depth \& Pose Supervision}
The key supervision signal for our depth and pose training comes from the task of novel view synthesis: given one input view of a scene, synthesize a new image of the scene seen from a different camera pose. We can synthesize a target view given a per-pixel depth in that image, plus the pose and visibility in a nearby view.\par
Let us denote the two images in each training sequence by $I_t$ (target) and $I_s$ (source). The view synthesis objective here can be formulated as
\begin{equation} \label{eq:1}
    \mathcal{L}_{vs} = \sum_p |I_t(p) - \hat{I_s}(p)|
\end{equation}
where $p$ indexes over pixel coordinates, and $\hat{I}_s$ is the source view $I_s$ warped to the target coordinate frame. (This view synthesis objective can be further applied to a larger sequence of images with one target image and multiple source images.)\par
The target view $I_t$ in the aforementioned equation is synthesized by a depth-based image renderer, that uses the source image $I_s$, the predicted depth map $\hat{D}_t$ and the estimated relative camera transformation matrix (pose) $\hat{T}_{t\rightarrow s}$, to generate the image from a novel viewpoint. Let $p_t$ denote the homogeneous coordinates of a pixel in the target view, and K denote the camera intrinsics matrix. We can obtain $p_t$’s projected coordinates onto the source view $p_s$ by
\begin{equation} \label{eq:2}
    p_s \sim K\hat{T}_{t\rightarrow s}\hat{D}_t(p_t)K^{-1}p_t
\end{equation}

The projected coordinates $p_s$ are continuous in nature. Hence, we use bilinear sampling of the four neighbouring coordinates to compute the values of the synthesized image $\hat{I_s(p_t)}$. (For more details on the view synthesis formulation, please refer \cite{zhou2017unsupervised}.)

\subsubsection{Optical Flow Supervision}
Similar to the novel view synthesis formulation above, we use image reconstruction using optical flow as our supervision method for training the FD. Using the same notation from above, the image reconstruction objective can be written as
\begin{equation} \label{eq:3}
    \mathcal{L}_{ir} = \sum_p |I_s(p) - \hat{I_t}(p)|
\end{equation}
 Here, $\hat{I}_t$ is the warped target image obtained after applying the predicted optical flow field to the source image.

\subsection{Objective Function}
\subsubsection{View Synthesis \& Image Reconstruction Loss}
The novel view synthesis and image reconstruction formulations mentioned in Eq. \ref{eq:1} and Eq. \ref{eq:3} are computed using the generalized Charbonnier loss function
\begin{equation} \label{eq:4}
    g(I, \hat{I}) = [\beta(I-\hat{I})^2 + \epsilon]^\alpha
\end{equation}
where, $I$ and $\hat{I}$ represent the reference and synthesize/warped images respectively. $\alpha$, $\beta$ are parameters for the Charbonnier loss and are set to: $\alpha = 0.45$, $\beta = 1$. $\epsilon = 10^{-7}$ is the fuzz factor used to avoid zeros.

\subsubsection{Smoothness Loss}
For the depth predictions, to filter out erroneous predictions and preserve sharp details, we use an edge-aware, second order smoothness loss weighted by image gradients
\begin{equation} \label{eq:5}
    \mathcal{L}_{sd} = (|\nabla_x^2D|\cdot e^{-10|\nabla_xI|} + |\nabla_y^2D|\cdot e^{-10|\nabla_yI|}) / 2
\end{equation}
where $\nabla_x$ and $\nabla^2_x$ represent the first and second order derivatives in the $x$ axis respectively, D is the predicted depth map, I is the input source image and $|.|$ denotes the element-wise absolute value.

\subsubsection{Forward-Backward Consistency Check}
For the estimated pose, we employ a forward-backward consistency check to ensure our pose prediction is robust. This check is enabled by making two passes of the network on the input images: once with the images default sequence ($[I_t, I_{t+1}$), and second after flipping the input image sequence ($[I_{t+1}, I_t]$). This allows us to get two pose predictions. Ideally, these two predictions are the exact opposites of each other. This loss is computed by first converting the predicted 6-DoF pose vectors into $4\times 4$ matrices. Next, the product of the two matrices is taken and a Charbonnier loss is applied to compute the loss between the product and a $4\times 4$ size identity matrix
\begin{equation} \label{eq:6}
    \mathcal{L}_p = [(P_{t\rightarrow t+1}\cdot P_{t+1\rightarrow t} - I_4)^2 + \epsilon]^{0.45}
\end{equation}
where $\alpha = 0.45$, $P_{t\rightarrow t+1}$ and $P_{t+1\rightarrow t}$ are the predicted pose matrices and $I_4$ is the identity matrix of size $4\times 4$.

\subsubsection{Final Loss Function}
All the above loss components are combined by taking a weighted sum. The final loss function then becomes
\begin{equation} \label{eq:7}
    \mathcal{L} = \lambda_{vs}\mathcal{L}_{vs} + \lambda_{ir}\mathcal{L}_{ir} + \lambda_{sd}\mathcal{L}_{sd} + \lambda_p\mathcal{L}_p
\end{equation}
where $\lambda_{vs}$, $\lambda_{ir}$, $\lambda_{sd}$ and $\lambda_{p}$ are the weights for view synthesis (depth), image reconstruction (flow), smoothness (depth) and pose loss components respectively.

\section{Training Details}
In this section, we firstly describe the implementation details of our network, followed by the datasets used for training.

\subsection{Implementation}
Our network is implemented in Python 3.7 using TensorFlow 2.0 library. The entire network is trained end-to-end jointly, using a single loss function.\par
The network is optimized by Adam, where $\beta_1$ = 0.9, $\beta_2$ = 0.999. Our initial learning rate is set to 0.0005 with an exponential decay with decay rate $=0.9$. We use a mini-batch size of 16. The loss weights are carefully selected after various experiments and set as: $\lambda_{vs}=1$, $\lambda_{ir}=1$, $\lambda_{p}=1$. $\lambda_{sd}=0$ for the first 50\% of the epochs and then increased to 1 for the next 50\%. The training process usually takes around 50 epochs to converge.\par
In order to speed-up training, we employed TensorFlow's mirrored strategy for training on multiple GPUs (all NVIDIA\textsuperscript{\textregistered} GeForce\textsuperscript{\textregistered} RTX 2080). In our experiments, we observed a 7x speed-up when training on 8 GPUs as compared to training on one GPU.

\subsection{Datasets Used}
We structured our input data pipeline to handle training on multiple datasets for easier experimentation. We first trained our network using the Cityscapes dataset ($\sim$80,000 images), followed by another round of training on KITTI ($\sim$40,000 images). The Cityscapes dataset offers a larger data corpus in the same domain as KITTI (driving dataset)\footnote{Training the network with a mix of the two datasets is not possible due to the difference in the camera characteristics}. In order to make our network more robust to more complicated scene structures, we also tried fine-tuning our network on Sintel dataset\footnote{Fine-tuning on Sintel did not yield any coherent results, possibly due to the small size of the dataset ($\sim$1100 images)}.\par
We resize the training image sequences to $192\times608$. Various image augmentation techniques, like random photometric adjustments (brightness, contrast, hue, saturation) were applied to the images during training. Moreover, random flipping and rotation (by 90, 180 and 270 degrees) was also applied to avoid overfitting.

\subsection{Hyperparameters}
We use 5 pyramid levels for our feature extraction encoder. Since our decoder heads are also pyramidal in nature, and make use of the raw features from the encoder stage, we end up with 5 levels convolutions for each of the decoder heads as well. After rigorous experimentation and tuning the number and of convolution filters at each level of the encoder and decoder heads, we arrived at a final model size of $\sim$8.4 million parameters, which is less than 5\% of the total size of state-of-the-art models in the same domain (unsupervised and monocular, depth, flow and pose estimation) \cite{yin2018geonet}.

\begin{table*}[t]
\centering
\begin{tabular}{l|c|c|c|c|c|c|c}
\hline
              & Dataset & Model Size & Abs Rel $\downarrow$ & Sq Rel $\downarrow$ & RMSE $\downarrow$  & RMSE log $\downarrow$ & Acc ($\delta$ \textless 1.25) $\uparrow$ \\ \hline
Eigen et al. \cite{eigen2014depth}  & K       & -          & 0.203   & 1.548  & 6.307 & 0.282    & 0.702                 \\ \hline
Zhou et al. \cite{zhou2017unsupervised}  & K       & 33.2M      & 0.183   & 1.595  & 6.709 & 0.270    & 0.734                 \\ \hline
Godard et al. \cite{godard2017unsupervised} & K       & 14.8M      & 0.115   & 0.882  & 4.701 & 0.190    & 0.879                 \\ \hline
Yang et al. \cite{yang2020d3vo}  & K       & -          & 0.099   & 0.763  & 4.485 & 0.185    & 0.885                 \\ \hline
Yin et al. \cite{yin2018geonet}   & CS+K    & 178.6M     & 0.153   & 1.328  & 5.737 & 0.232    & 0.802                 \\ \hline
Casser et al. \cite{casser2019depth} & K       & -          & 0.109   & 0.825  & 4.750 & 0.187    & 0.874                 \\ \hline
DFPNet (Ours) & CS+K    & 8.4M       & 0.257   & 3.107  & 7.845 & 0.326    & 0.651                 \\ \hline
\end{tabular}
\caption{Monocular depth results on KITTI dataset, by the split of Eigen et al. \cite{eigen2014depth}. K refers to KITTI and CS refers to Cityscapes as the training dataset. The model size is the total number of trainable parameters for the particular method and has been computed from the respective author's provided model code.}
\label{table:1}
\end{table*}

\section{Results}
The qualitative and quantitative results in monocular depth, optical flow and camera pose estimation tasks are discussed in this section. A visualization of the qualitative results is shown in Figure \ref{fig:picture1}.

\subsection{Monocular Depth Estimation}
We evaluate our model for depth estimation on KITTI \cite{geiger2013vision} and Cityscapes \cite{cordts2015cityscapes} dataset and compare its performance on some of the recent state-of-the-art depth estimation methods - both supervised and unsupervised. The models are compared against 5 standard KPIs: Absolute Relative Error (Abs Rel), Square Relative Error (Sq Rel), RMSE, log RMSE and Accuracy with threshold 1.25.\par
For KITTI evaluation, we use the test split provided by Eigen et al. \cite{eigen2014depth} to compare with other related works. We exclude all the static scenes from the test set. During evaluation, we resize our outputs to the ground truth image resolution ($512\times1382$). We follow the evaluation method used by Godard et al. \cite{godard2019digging} for a consistent comparison. A comparison of results for KITTI evaluation are shown in Table \ref{table:1}.\par
For Cityscapes evaluation, we use the built-in test split of the dataset. There is not much literature on Cityscapes evaluation on depth estimation. Consequently, unlike KITTI, there is no standard evaluation protocol defined on this dataset. So, we use the same evaluation method as KITTI. Table \ref{table:2} shows the results of evaluation on Cityscapes.

\begin{table*}[t]
\centering
\begin{tabular}{l|c|c|c|c|c|c}
\hline
              & Dataset & Abs Rel $\downarrow$ & Sq Rel $\downarrow$ & RMSE $\downarrow$  & RMSE log $\downarrow$ & Acc ($\delta$ \textless 1.25) $\uparrow$ \\ \hline
Pilzer et al. \cite{pilzer2018unsupervised} & K       & 0.440   & 6.036  & 5.443  & 0.398    & 0.730                 \\ \hline
Casser et al. \cite{casser2019unsupervised} & CS      & 0.151   & 2.492  & 7.024  & 0.202    & 0.826                 \\ \hline
Li et al. \cite{li2020unsupervised}    & CS+K    & 0.119   & 1.290  & 6.980  & 0.190    & 0.846                 \\ \hline
DFPNet (Ours) & CS      & 0.228   & 3.169  & 10.417 & 0.304    & 0.633                 \\ \hline
\end{tabular}
\caption{Monocular depth results on Cityscapes dataset.}
\label{table:2}
\end{table*}

\subsection{Optical Flow Estimation}
The performance of optical flow component is validated on the KITTI stereo/flow test split and Sintel \cite{mayer2016large} clean pass test set. Again, we compare the performance of our model against both, supervised and unsupervised state-of-the-art methods. The KPIs reported are End Point Error (EPE) and the Error Rate (ER). Table \ref{table:3} summarizes the results on KITTI and Table \ref{table:4} summarizes the results on Sintel.\par

\begin{table*}[h!]
\centering
\begin{tabular}{l|c|c|c}
\hline
                    & Size & EPE $\downarrow$   & ER (\%) $\downarrow$ \\ \hline
Teed et al. \cite{teed2020raft}         & 5.3M       & 0.63  & 1.50    \\ \hline
Sun et al. \cite{sun2018pwc}          & 9.4M       & 2.16  & 9.80    \\ \hline
Jonschkowski et al. \cite{jonschkowski2020matters} & 5.7M       & 2.71  & 9.05    \\ \hline
Yin et al. \cite{yin2018geonet}          & 178.6M     & 10.81 & -       \\ \hline
DFPNet (Ours)       & 8.4M       & 3.86  & 10.17   \\ \hline
\end{tabular}
\caption{Optical flow evaluation results on the KITTI flow split.}
\label{table:3}
\end{table*}

\subsection{Camera Pose Estimation}
To evaluate the performance of our pose estimation network, we ran our network on the official KITTI odometry split. This split containing 11 driving sequences with ground truth odometry obtained through the IMU/GPS readings, which we use for evaluation. For comparison with other models, we report the results on sequences 09 and 10 in Table \ref{table:5}. Unlike others, we do not fine-tune our network on the rest of the sequences in the KITTI odometry split (sequences 00-08).

\begin{table}[h!]
\centering
\begin{tabular}{l|c|c}
\hline
              & Seq 09                     & Seq 10 \\ \hline
Zhou et al. \cite{zhou2017unsupervised}   & 0.016                      & 0.013  \\ \hline
Godard et al. \cite{godard2019digging} & 0.017                      & 0.015  \\ \hline
Yin et al. \cite{yin2018geonet}    & 0.012                      & 0.012  \\ \hline
DFPNet (Ours) & \multicolumn{1}{l|}{0.065} & 0.043  \\ \hline
\end{tabular}
\caption{ATE (absolute trajectory error) on KITTI Odometry split.}
\label{table:5}
\end{table}

\section{Conclusion}
We proposed DFPNet, a framework for unsupervised, end-to-end, joint learning of monocular depth, optical flow and camera pose prediction. Our pyramidal feature extraction technique and unsupervised training paradigm, profoundly reveals the capability of neural networks in capturing both high level cues and feature correspondences for capturing multiview geometry characteristics. Our network structure, paired with the composite loss function, demonstrates how neural networks can accomplish complicated vision tasks with seemingly small number of parameters, (compared to other state-of-the-art methods). For future work, we would like to incorporate the use of geometric consistency between depth map and optical flow field for more coherency between the predictions. We also plan to design and implement a hardware accelerator for the proposed model for optimized, real-time performance.

\begin{table}[t]
\centering
\begin{tabular}{l|c}
\hline
                    & EPE $\downarrow$ \\ \hline
Teed et al. \cite{teed2020raft}         & 0.76 \\ \hline
Sun et al. \cite{sun2018pwc}         & 2.50 \\ \hline
Jonschkowski et al. \cite{jonschkowski2020matters} & 2.02 \\ \hline
DFPNet (Ours)       & 4.48 \\ \hline
\end{tabular}
\caption{Optical flow evaluation results on Sintel (train, clean split)}
\label{table:4}
\end{table}

{\small
\bibliographystyle{ieee_fullname}
\bibliography{egbib}
}

\end{document}